%% file: iclr2025_arXiv.tex
\documentclass{article} % For LaTeX2e
\usepackage{iclr2025_conference,times}

\input{math_commands.tex}

\usepackage{hyperref}      
\usepackage{xcolor}

\hypersetup{
    colorlinks=true,    
    linkcolor=brown,      
    urlcolor=brown,   
    citecolor=brown       
}
\usepackage{url}
\usepackage{booktabs}
\usepackage{multicol}
\usepackage{multirow}
\usepackage{graphicx}
\usepackage{wrapfig}
\usepackage{enumitem} % chen add
\usepackage{xcolor}
\usepackage{colortbl}

\newcommand{\approach}{\text{\textit{HandsOnVLM}}}

\newcommand{\rebuttalcolortext}[1]{{#1}}

\title{HandsOnVLM: Vision-Language Models for Hand-Object Interaction Prediction}

\author{Chen Bao$^1$, Jiarui Xu$^2$, Xiaolong Wang$^{2,\dag}$, Abhinav Gupta$^{1,\dag}$, Homanga Bharadhwaj$^{1,}$\thanks{denotes equal advising. Correspondence to \texttt{chenbao@cmu.edu, hbharadh@cs.cmu.edu}}\\
$^{1}$ The Robotics Institute, Carnegie Mellon University \\
$^{2}$ University of California at San Diego
}

\iclrfinalcopy % Uncomment for camera-ready version, but NOT for submission.
\begin{document}

\maketitle

\begin{abstract}
How can we predict future interaction trajectories of human hands in a scene given high-level colloquial task specifications in the form of natural language? In this paper, we extend the classic hand trajectory prediction task to two tasks involving explicit or implicit language queries. Our proposed tasks require extensive understanding of human daily activities and reasoning abilities about what is happening next given cues from the current scene. We also develop new benchmarks to evaluate the proposed two tasks, Vanilla Hand Prediction (VHP) and Reasoning-Based Hand Prediction (RBHP). We enable solving these tasks by integrating high-level world knowledge and reasoning capabilities of Vision-Language Models (VLMs) with the auto-regressive nature of low-level ego-centric hand trajectories. Our model,~\approach~is a novel VLM that can generate textual responses and produce future hand trajectories through natural-language conversations. Our experiments show that~\approach~outperforms existing task-specific methods and other VLM baselines on proposed tasks, and demonstrates its ability to effectively utilize world knowledge for reasoning about low-level human hand trajectories based on the provided context. More details can
be found at \url{https://www.chenbao.tech/handsonvlm/}.
\end{abstract}

\section{Introduction}
\begin{figure}[h!]
\includegraphics[width=1\textwidth,trim={0 7.5cm 0 0},clip]{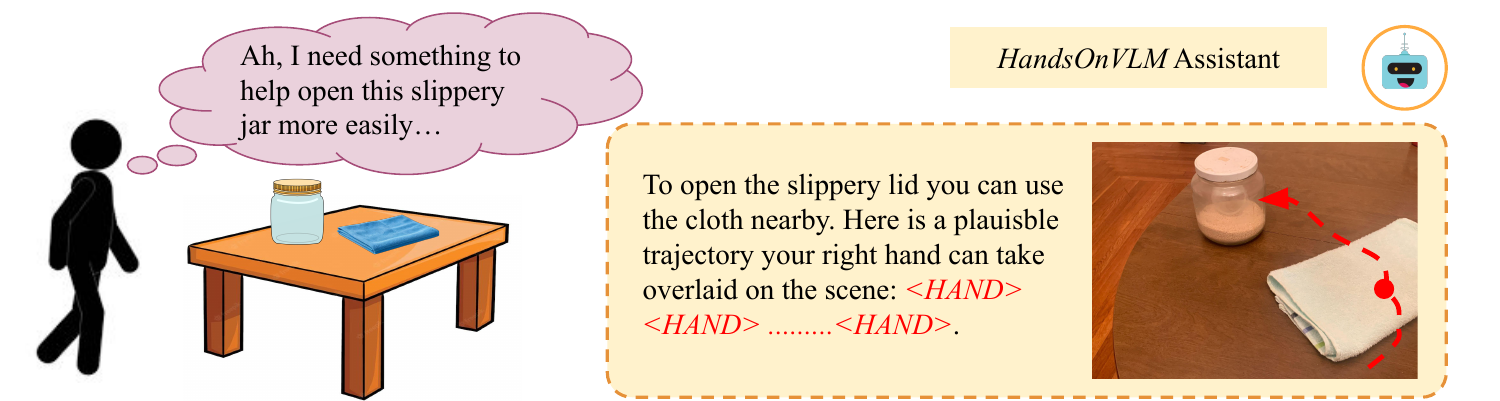}
\caption{\approach~forecasts low-level actions in the form of hand trajectories in the user's egocentric view of a scene when queried with a question via natural language.}
\label{fig:teaser}
\end{figure}

Humans interact with the everyday world and express themselves with informal and oftentimes vague language descriptions. Consider the example in Fig.~\ref{fig:teaser} - when we try to open the jar, we might think, ``Ah, I need something to help open this slippery jar more easily." We are uncertain about \textit{what} we exactly want as well as about \textit{how} to come up with a solution. To build a computational system for addressing this need, we would require a good understanding of what tools we have lying around (visual scene understanding), general apriori experience of opening jars (reasoning ability and world knowledge priors), and the ability to actually execute the necessary actions for opening the jar (low-level trajectory). In this paper, we develop two language-conditioned tasks for tackling this problem, propose benchmarks for evaluating progress on these tasks, and build a vision-language model (VLM) for predicting low-level hand trajectories in a user's egocentric view of a scene given colloquial language queries.

Towards a similar goal, some prior works have focused on identifying human intentions based on egocentric human videos of daily activities (high-level intentions of the form ``cutting pepper", ``washing plates")~\citep{krishna2017densecaptioningeventsvideos, grauman2022ego4dworld3000hours, kahatapitiya2024victrvideoconditionedtextrepresentations}, while others have focused on predicting low-level actions such as hand trajectories given human action clips~\citep{liu2022jointhandmotioninteraction,zhang2024pear} without conditioning the predictions on detailed language descriptions of the task to be performed. Both these scenarios are a bit restrictive since for most everyday tasks (e.g. in Fig.~\ref{fig:teaser}) we need a combination of high-level reasoning of what to do in a scene and low-level understanding of how to interact with the relevant objects in the scene.

By drawing on the recent successes of VLMs for high-level reasoning~\citep{liu2023visualinstructiontuningllava, lai2024lisareasoningsegmentationlarge, cheng2024spatialrgptgroundedspatialreasoning} and advancements in hand reconstructions from generic web videos~\citep{shan2020handobject,frankmocap,pavlakos2023reconstructinghands3dtransformers}, we develop a system for future hand trajectory prediction given conversation-style language instructions. Current best multimodal VLMs are good at predicting semantic actions in the form of \textit{what} is happening at a certain point in a video (\cite{maaz2023videochatgpt, huang2024litalanguageinstructedtemporallocalization}), interpreting what objects are in a scene~\citep{openai2024gpt4technicalreport} and natively support free-form language conversations for conditioning. However, they are not good at directly predicting \textit{low-level} actions (in the future) of the form of hand-object trajectories. At the same time, recovering low-level interactions in videos, like hand meshes~\citep{pavlakos2023reconstructinghands3dtransformers}, object meshes \citep{HOLD}, and regions of interactions~\citep{shan2020handobject,goyal2022human} has independently become very reliable in recent years. Our key insight is to fine-tune a pre-trained VLM with auto-regressive trajectory predictions of human hand positions, given a few seconds of video and a language description of the task.

Our approach~\approach~casts hand trajectory prediction as an auto-regressive next token prediction conditioned on fused video and language tokens. We develop~\approach~as an interactive chat assistant that we can query with informal instructions of the form, ``Where should my hand move if I want to open the refrigerator?'' and a video (or an image) of a scene, and obtain outputs of the form, ``To open the refrigerator, the predicted hand trajectory is~\textless\textit{HAND}\textgreater~,....~\textless\textit{HAND}\textgreater~'' The~\approach~model first converts the RGB video context to visual tokens and fuses them with the language tokens through slow-fast pooling~\citep{huang2024litalanguageinstructedtemporallocalization} for capturing temporal information from the context video at a fine resolution. We extend the vocabulary to add a new~\textless\textit{HAND}\textgreater~token, and output a sequence of text and and hand tokens. We finally have a trajectory decoder to convert the hand tokens to a sequence of 2D positions of the left and right hands over the prediction horizon. 

In summary, our paper has the following contributions:
\begin{itemize}
\item We develop~\approach, a novel VLM that can generate textual responses and produce future hand trajectories through conversations by expanding the original vocabulary with hand tokens and having iterative position encodings for auto-regressive predictions during inference. 
\item We extend existing traditional hand prediction tasks to two new tasks, Vanilla Hand Prediction (VHP) and Reasoning-based Hand Prediction (RBHP), to predict hand trajectories from ego-centric human videos conditioned on language queries of different forms.
\item We develop benchmarks for evaluating progress on the VHP and RBHP tasks which we will open-source to the community, in addition to our trained models on the benchmarks.
\end{itemize}

Our results on diverse real-world datasets of human videos and zero-shot evaluations on completely unseen datasets demonstrate strong generalization and reasoning capabilities of~\approach~for hand trajectory prediction given colloquial language instructions. Furthermore, the model outperforms most baselines on the Reasoning-based Hand Prediction (RBHP) task, showcasing its capability to reason and leverage world knowledge of VLMs.

\section{Related Work}
We discuss prior works on human motion reconstruction and forecasting, developments in multimodal large language models and action understanding from human videos. 

\subsection{Human Motion Reconstruction and Forecasting}

Several prior works have attempted to recover hand meshes and full body meshes from human videos~\citep{frankmocap,pavlakos2023reconstructinghands3dtransformers}. Going beyond reconstruction, other works have also investigated forecasting motions of humans in the future. Early works used RNNs~\citep{DeepRepresentationLearningforHumanMotionPredictionandClassification, bütepage2017anticipatingfuturesonlinehuman, DBLP:conf/bmvc/HondaKN20} for anticipating future human poses, and recent approaches include Transformer architectures for more diverse and plausible future predictions~\cite{ding2023expressive}. More directly related to our work, some approaches predict egocentric hand-trajectories in the form of 2D waypoints~\citep{liu2020fhoi}, and others also predict object affordances jointly with hand trajectories~\citep{liu2022jointhandmotioninteraction, bharadhwaj2024towards}. Some predict hand trajectories in a 3D space conditioned on a few RGB observations from an egocentric view~\citep{bao2023uncertaintyawarestatespacetransformer}. Architectures for such egocentric predictions have ranged from transformers~\citep{liu2022jointhandmotioninteraction,bao2023uncertaintyawarestatespacetransformer} to diffusion models~\citep{ma2024diffip2ddiffusionbasedhandobjectinteraction,ma2024madiffmotionawaremambadiffusion} trained specifically for this prediction task. Our work extends this line of low-level egocentric trajectory prediction by enabling reasoning capabilities through augmentation and joint training with a pre-trained VLM.

\subsection{Multimodal Large Language Models}
 Our work is enabled by developments in multimodal Large Language Models that augment vision and language reasoning in a unified model. Such models like LLaVA ~\citep{liu2023visualinstructiontuningllava} and Video-ChatGPT~\citep{maaz2023videochatgpt} have enabled large-scale video understanding and localization of temporal events (semantic actions) in videos~\citep{huang2024litalanguageinstructedtemporallocalization}. Adjacently, other works have sought to make the inputs to the VLMs more flexible and informal through automatic segmentations of language instructions~\citep{lai2024lisareasoningsegmentationlarge,yang2024lisaplus} and visual grounding allowing flexibility to process both image and region inputs~\citep{hanoona2023GLaMM}. Recent works have extended the capabilities of VLMs to diverse domains including robotic navigation~\citep{zhang2024navidvideobasedvlmplans}, robotic manipulation~\citep{kim2024openvla,rt22023arxiv}, spatial reasoning~\cite{cheng2024spatialrgptgroundedspatialreasoning}, and reasoning about 3D human poses from images and text~\cite{feng2024chatpose}. While these approaches are orthogonal to our task of egocentric hand trajectory prediction, they serve as evidence of the potential of VLMs for downstream applications.

\subsection{Action Recognition and Prediction from Videos}
Understanding actions in the form of what is happening in a video segment has a long history in computer vision~\citep{actionrecognition2,actionrecognition1,2010humanactionrecognition,feichtenhofer2019slowfast}. Several benchmarks and datasets containing human videos and action labels for tasks have also been proposed for related problems~\citep{grauman2022ego4dworld3000hours,caba2015activitynet,goyal2017something, pmlr-v205-xiong23a}. Our work leverages such datasets and goes beyond \textit{recognition} of actions in videos to \textit{prediction} of low-level actions in the future by first reasoning about future high-level actions through a VLM. As such our work can have potential applications in robotics for learning motion from web videos for manipulation by complementing prior works in this space~\citep{bharadhwaj2024gen2act,bahl2023affordances,nair2022r3m,bharadhwaj2025track2act}.

\section{Approach}
\label{sec:approach}
\approach~is a video-based VLM with the capability of predicting future hand trajectories given a video context and language instructions. There are three key components of~\approach's architecture: (1) SlowFast tokens to capture temporal information at fine temporal resolution, (2) hand representation using an augmented vocabulary of~\textless\textit{HAND}\textgreater~token, and (3) iterative hand decoding to enable auto-regressive trajectory training and inference. In the training stage, we fine-tune a pre-trained VLM by combining next-token prediction loss and trajectory loss.

\begin{figure}[htbp!]
\includegraphics[width=1\textwidth,trim={0 7.15cm 0 0},clip]{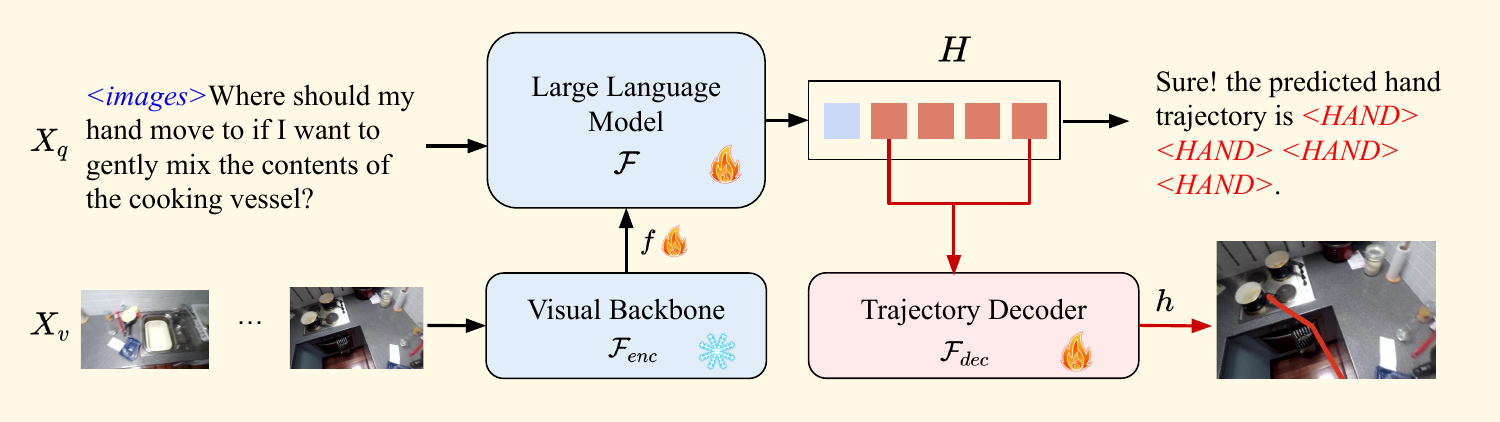}
\caption{Overview of the~\approach~architecture, where \includegraphics[height=1em]{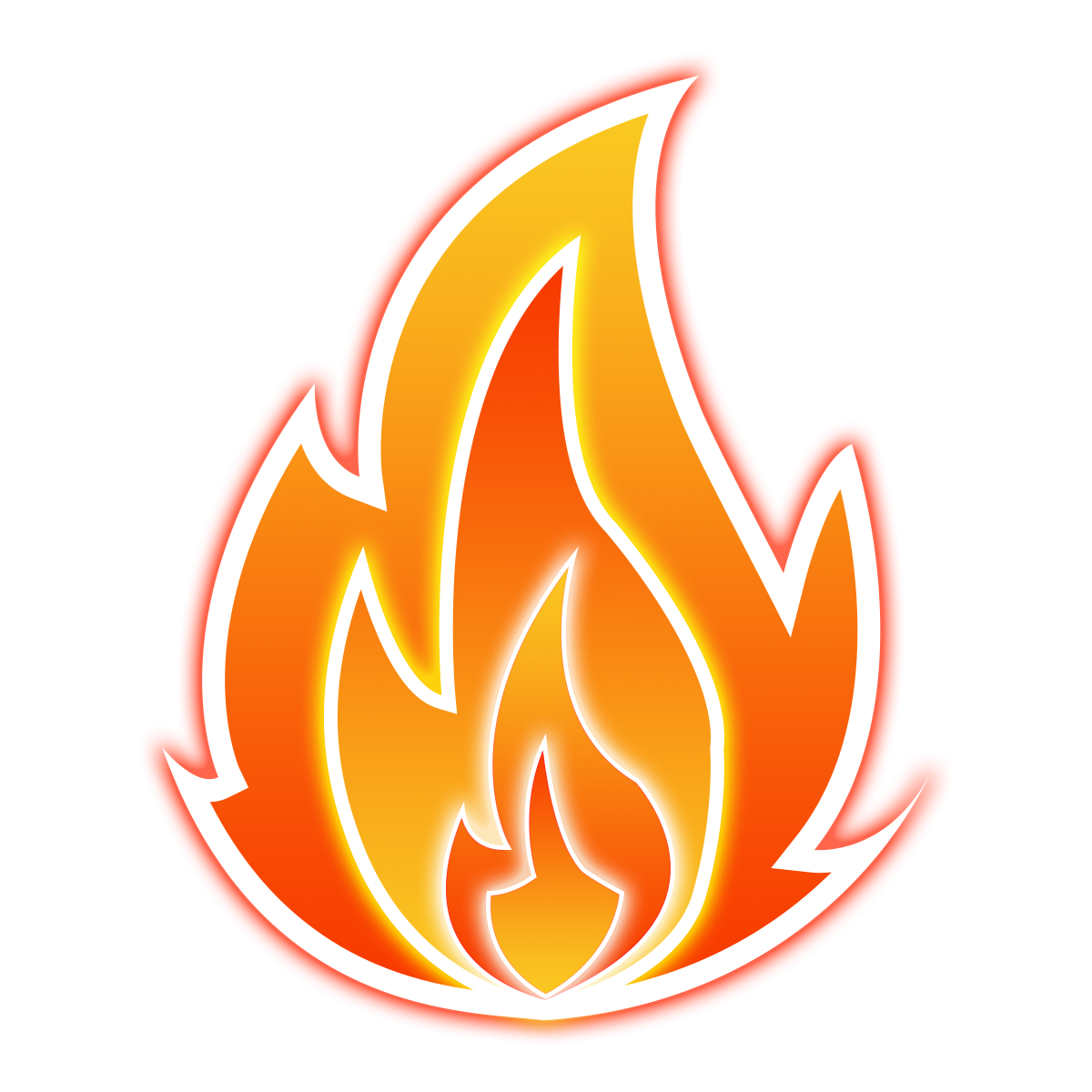} and \includegraphics[height=1em]{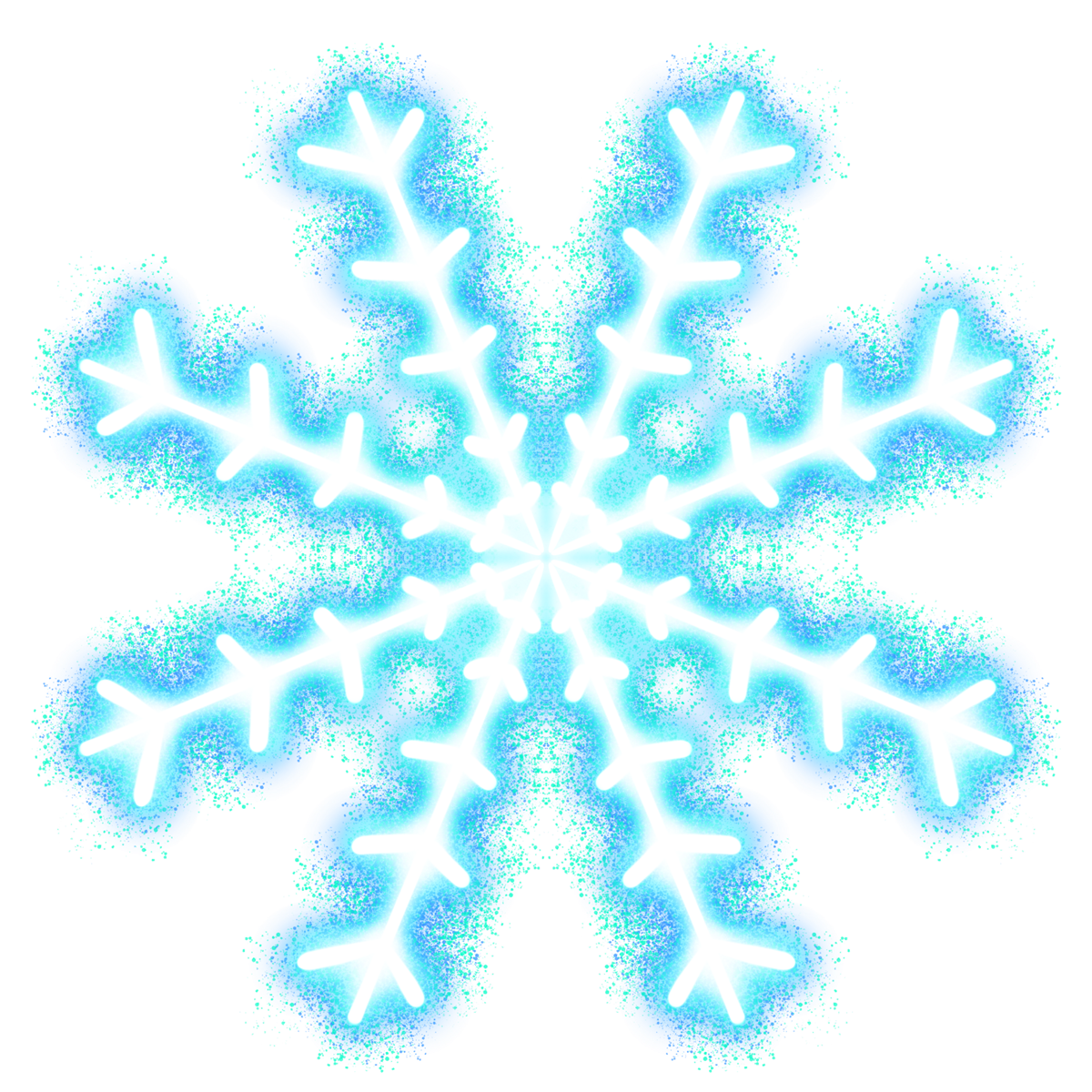} denote trainable and frozen modules separately. ~\approach~casts hand trajectory prediction as an auto-regressive next token prediction conditioned on fused video and language tokens. The architecture augments a pre-trained VLM with an additional hand token in the vocabulary. We use \includegraphics[height=1em]{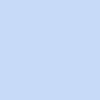} and \includegraphics[height=1em]{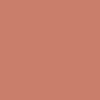} to represent text and~\textless\textit{HAND}\textgreater~tokens respectively. }
\label{fig:arch}
\vspace{-0.5cm}
\end{figure}

\subsection{Architecture}
We show an overview of the~\approach~model architecture in Fig~\ref{fig:arch}.\rebuttalcolortext{~\approach~takes a sequence of $T$ frames $X_v$ and a language instruction $X_q$ as input and predicts future hand trajectories $\mathcal{H}=\{h_{T+i}\}_1^N$, where $N$ is the future horizon. At each future time step $T+i$, the future hand location $h_{T+i}$ consists of the 2D location of the center of the left and right hands projected to the last observation frame $X_v\left[-1\right]$.} The key components of the architecture include a visual backbone $\mathcal{F}_{enc}$, a vision-to-language projection layer $f$, a Large Language Model(LLM) $\mathcal{F}$ and a trajectory decoder $\mathcal{F}_{dec}$.

\textbf{SlowFast Token Compression.} To obtain a capable video-conditioned VLM we need to be able to interpret temporal information at a fine resolution. Following ~\cite{huang2024litalanguageinstructedtemporallocalization}, given $X_v$, we embed them into $T \times M$ visual tokens using a visual backbone, where $M$ is the number of tokens in each frame. Then we apply slow-fast pooling to get $T+M$ visual tokens. In the fast path, we average all the tokens within each frame to get $T$ tokens overall. We also uniformly select $s$ frames among all $T$ frames and perform $s\times s$ spatial average pooling to get $M$ slow frames in total. These slow tokens will help preserve spatial information during the encoding process. \rebuttalcolortext{Then we embed and align $T+M$ visual tokens to the language space through a vision-to-language projector $f(\cdot)$.}

\textbf{Hand as Embedding.} To represent hand in the language space, we extend the existing vocabulary with a new~\textless\textit{HAND}\textgreater~token. \rebuttalcolortext{However, a typical embedding layer would encode each~\textless\textit{HAND}\textgreater~token identically, resulting in individual~\textless\textit{HAND}\textgreater~token being indistinguishable from one another. To overcome this limitation, we embed ground truth hand positions into the~\textless\textit{HAND}\textgreater~tokens during the tokenization process. We feed them into the Large Language Model backbone and get the embedding of the last layer $H$, where $H=\mathcal{F}(X_q, f(\mathcal{F}_{enc}(X_v)))$}.

\textbf{Iterative Hand Decoding.} \rebuttalcolortext{For $i$-th token in the sequence, let $H_i$ be the last-layer embedding of this token from the Large Language Model.~\approach~decode it to predict the $(i+1)$-th token as LLMs do. When $(i+1)$-th token is a~\textless\textit{HAND}\textgreater~token, we input $H_i$ into a hand trajectory decoder $\mathcal{F}_{dec}$ to predict the hand position of the $(i+1)$-th token $h_{i+1}=\mathcal{F}_{dec}(H_i)$. During inference, this decoded position is then encoded into the corresponding~\textless\textit{HAND}\textgreater~token embedding for following prediction rounds. In this way, we ensure that each subsequent prediction is conditioned on all previously predicted hand positions, maintaining temporal consistency and spatial awareness throughout the inference process and mitigating compounding errors.}

\subsection{Training Objectives}
The model is trained end-to-end using a text generation loss $\mathcal{L}_{\text{txt}}$ and a hand trajectory prediction loss $\mathcal{L}_{\text{hand}}$. The overall objective $\mathcal{L}$ is the weighted sum of both losses, determined by $\lambda_{t x t}$ and $\lambda_{\text {hand}}$:

\begin{equation}
\mathcal{L} = \lambda_{\text{txt}} \mathcal{L}_{\text{txt}} + \lambda_{\text{hand}} \mathcal{L}_{\text{hand}}
\end{equation}

Specifically, $\mathcal{L}_{\text{txt}}$ is the auto-regressive cross-entropy loss
for text generation, and $\mathcal{L}_{\text {hand}}$ is the hand prediction loss, which encourages the model to generate high-quality hand trajectories as well. Following \cite{liu2022jointhandmotioninteraction}, we employ a reconstruction loss over future timesteps and a KL-Divergence Regularization loss as $\mathcal{L}_{\text {hand}}$:
\begin{equation}
\mathcal{L}_{\text {hand}}=\sum_{t=1}^N \mathcal{L}_{\text {recon }}\left(h_{T+t}, \hat{h}_{T+t}\right)+\mathcal{L}_{k l}\left(\mu_h, \sigma_h\right).
\end{equation}
We employ CVAE \citep{cvae} as the hand trajectory decoder in this work (although the method is not tied to it). Thus, $\mathcal{L}_{\text {recon }}$ is the MSE loss over valid hand positions, and $\mu_h$, $\sigma_h$ here are the mean and the standard deviation that regularizes the latent z-space to be close to the normal distribution.

\section{Reasoning and Predicting Hand Trajectories}
In this section, we introduce two tasks: the Vanilla Hand Prediction (VHP) task, which extends the classic hand motion prediction \citep{liu2022jointhandmotioninteraction}, and the proposed Reasoning-based Hand Prediction (RBHP) task. Finally, we describe a two-step annotation-generating pipeline to build the corresponding RBHP dataset.

\subsection{Vanilla Hand Prediction Task}
\label{sec:dataset generation process}
In this task, explicit action narration is required to predict the next hand motion. Here explicit means the action narration directly specifies the action and the target object without ambiguity, such as ``cut the paper" or ``open the microwave". We choose Epic-Kitchen \citep{ek55, Damen2022EPIC100}, H2O \citep{h2odataset} and FPHA \citep{garciahernando2018fpha} as datasets for this task. \rebuttalcolortext{To generate the hand labels for all the datasets, following \citep{liu2022jointhandmotioninteraction}, we first run an off-the-shelf active hand-object detector \citep{handobjectdetector} to get the hand bounding box in each frame. To get the ground truth of each future hand trajectory, we first compute pairwise homographies by matching SURF \citep{bay2006surf} features of masked regions through RANSAC and project each future hand position into the last observation frame. Then, we apply cubic Hermite spline interpolation to smooth the projected trajectories and fill any missing points. Finally, we filter the resulting trajectories with multiple criteria, including confidence thresholds, highest-score detection selection, feature matching thresholds, trajectory completeness checks, and boundary constraints.
}

To reformat these datasets for visual question answering, we structure them in a question-answer format using the following template: 

\textit{``USER:}\textless \textit{images}\textgreater \textit{,can you give me the future hand trajectory for \{explicit action narration\}?}$\textit{ASSISTANT: Sure, it is}$\textless\textit{HAND}\textgreater\textless\textit{HAND}\textgreater\textless\textit{HAND}\textgreater\textless\textit{HAND}\textgreater\textit{."},

where \textless \textit{images}\textgreater represents a placeholder of visual tokens of the input frames. Note that the action is optional because we can also generate general templates without specifying the action, and in this case the task reduces to that in prior works~\citep{liu2022jointhandmotioninteraction,bao2023uncertaintyawarestatespacetransformer, ma2024diffip2ddiffusionbasedhandobjectinteraction}.

\subsection{Reasoning-based Hand Prediction Task}
\label{sec:reasoning data annotation pipeline}
In addition to the Vanilla Hand Prediction Task, we introduce the Reasoning-based Hand Prediction (RBHP) task. Instead of utilizing explicit instructions to directly predict the hand motion, here the system is required to reason about it with implicit instructions. We define implicit instructions as colloquial language instructions that provide sufficient information for inferring the intended human hand action through reasoning, without explicitly naming the target object or action.

\begin{figure}[htbp!]
  \centering
  \includegraphics[width=1.05\textwidth,trim={0 2.5cm 0 0},clip]{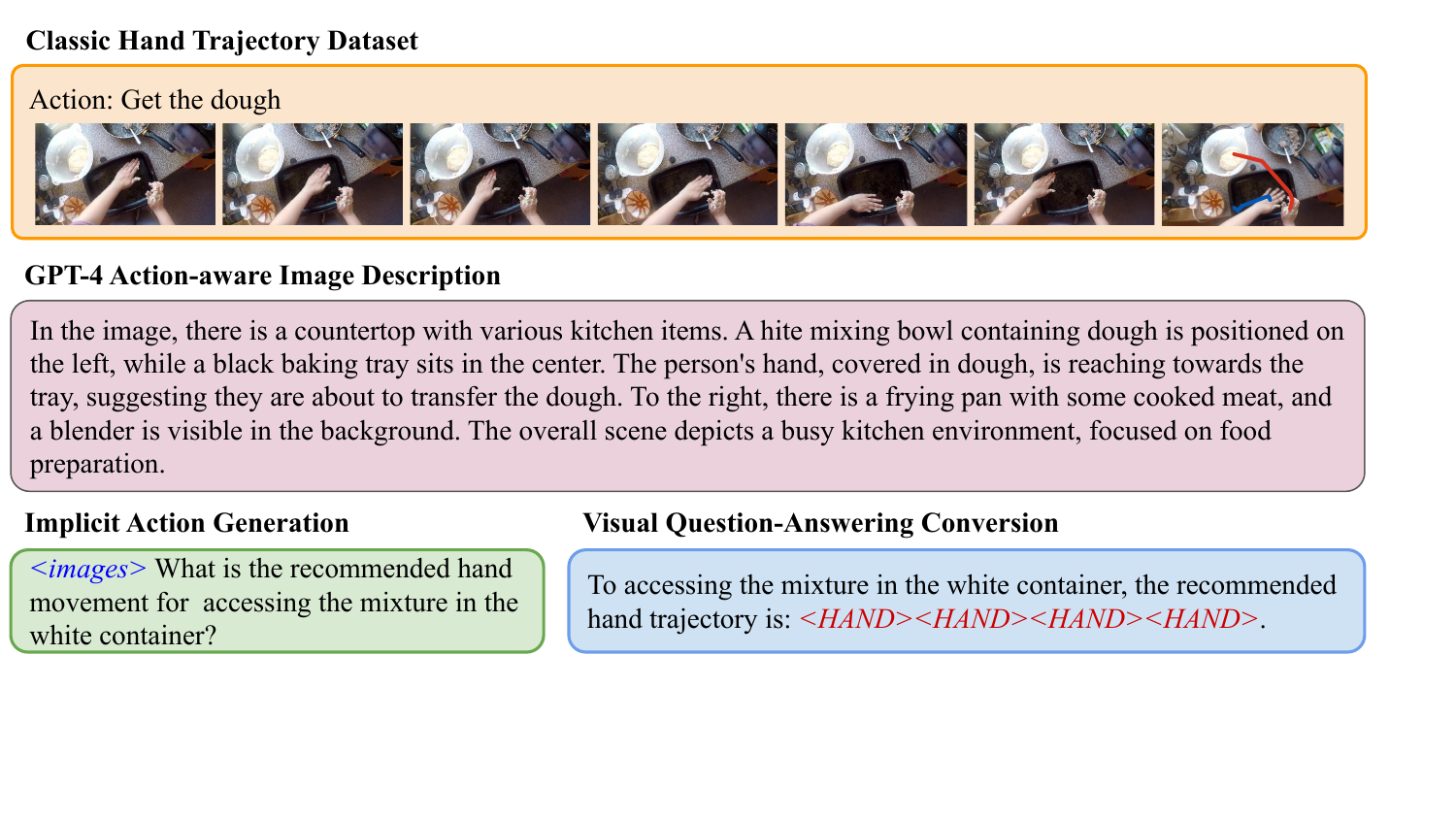}
  \caption{Illustration of the annotation pipeline for the RBHP task. By using GPT-4 on human video datasets we extract implicit language instructions for visual question-answering. The red and blue lines respectively show trajectories for the right and left hands.}
  \label{fig:dataset}
\end{figure}

To construct a dataset for this task, we implement a two-step annotation-generating pipeline (Fig. \ref{fig:dataset}) powered by GPT-4 \citep{openai2024gpt4technicalreport}. This pipeline extracts implicit instructions from the Epic-Kitchens-100 dataset \citep{Damen2022EPIC100}. Prompt templates for these two steps are provided in the Appendix \ref{sec: prompts}.

\textbf{Action-aware Image Description.} To get the implicit instructions, the first step is to generate a detailed description of the scene including all the objects in the foreground. We prompt GPT-4 with the ground truth action to capture action-related information, such as the physical properties of the target object or the spatial relationship with other objects.

\textbf{Implicit Action Generation.} Using the action-aware description of the scene, we are able to generate the implicit instructions using GPT-4 in a text template as follows: 

\textit{"USER:}\textless \textit{images}\textgreater \textit{,can you give me the future hand trajectory for \{action implicit description\}?}$\textit{ASSISTANT: Sure, it is}$\textless\textit{HAND}\textgreater\textless\textit{HAND}\textgreater\textless\textit{HAND}\textgreater\textless\textit{HAND}\textgreater\textit{."}.

\rebuttalcolortext{We choose Epic-Kitchen and Ego4D \citep{grauman2022ego4dworld3000hours} as datasets for this task. Through the annotation-generating pipeline, we generate 7.5k question-answering pairs from Epic-Kitchen, and 8k pairs from Ego4D for zero-shot evaluation.}

\section{Experiment}
We perform experiments for both the proposed tasks in order to answer the following research questions:
\begin{itemize}
    \item How plausible are the hand trajectories produced by~\approach?
    \item Does~\approach~exhibit reasoning abilities for implicit language queries?
    \item Does~\approach~generalize zero-shot to unseen scenes from new datasets?
\end{itemize}

\subsection{Experiment Details}
\label{sec:expdetails}
\textbf{Architecture.}  % copied from LITA
Following LITA's architecture, We use CLIP-L-14~\citep{radford2021learningclip} as the visual encoder and Vicuna~\citep{vicuna2023} as the LLM module. We adapt the vision-language projector from LLaVA~\citep{liu2023visualinstructiontuningllava} and have a CVAE~\citep{cvae} as trajectory decoder.

\textbf{Datasets.} For VHP and RBHP datasets, we sample 10 frames and predict the hand position in next 4 frames at $\text{FPS}=4$. \label{dataset} In addition to our proposed datasets,~\approach$^\dag$ are also trained on a few additional datasets for five different tasks, namely ActivityNet-Captions~\citep{krishna2017densecaptioningeventsvideos}and YouCook2~\citep{zhou2017automaticlearningproceduresweb} for dense video captioning and event localization, NExT-QA ~\citep{xiao2021next} for video question answering, LLaVA-150K~\citep{liu2023visualinstructiontuningllava} for image instruction tuning, ActivityNet-RTL~\citep{huang2024litalanguageinstructedtemporallocalization} for reasoning temporal localization. We co-train with these additional tasks to help with visual understanding and reasoning, and this is enabled by the flexible modeling of~\approach~that allows training on generic QA datasets. 

\textbf{Implementation Details.} For~\approach~and other VLM-based baselines, in each epoch we select 24K samples from the Epic-Kitchens-100 VHP dataset. For~\approach$^\dag$, in each epoch we randomly select 6K samples in Epic-Kitchens-100 VHP dataset, 6K in \rebuttalcolortext{Epic-Kitchens-100} RBHP dataset and another 12K that are uniformly distributed among all other 5 tasks. We use a batch size of 128, a learning rate of 2e-5 and train for 40 epochs. The total wall-clock time for training is around 18 hours for the 7B models while using 8 H100 GPUs. The LLM and vision-language projector are initialized with the LLaVA-1.3 pre-trained weights. During training, we freeze the visual backbone and fully fine-tune other modules.

\subsection{Metrics and Baselines}
\label{sec:baselines}
\begin{table*}[t]
\centering
\resizebox{\textwidth}{!}{%
\setlength\tabcolsep{2pt} 
\begin{tabular}{c|c|ccc|ccc|ccc|ccc}
\toprule
& &
\multicolumn{6}{c|}{On Validation Split}& 
\multicolumn{6}{c}{Zero-shot}
\\
\toprule
&
 &
\multicolumn{3}{c|}{EK55}& 
\multicolumn{3}{c|}{EK100}& 
\multicolumn{3}{c|}{H2O} & 
\multicolumn{3}{c}{FPHA}
\\

% \cline{6-11}
Approach & BBox Input & 
$\text {ADE} \downarrow$ &
$\text {FDE} \downarrow$ &
$\text {WDE} \downarrow$ &
$\text {ADE} \downarrow$ &
$\text {FDE} \downarrow$ &
$\text {WDE} \downarrow$ &
$\text {ADE} \downarrow$ &
$\text {FDE} \downarrow$ &
$\text {WDE} \downarrow$ &
$\text {ADE} \downarrow$ &
$\text {FDE} \downarrow$ &
$\text {WDE} \downarrow$
\\
\hline
KF & \checkmark & 0.392
 & 0.386 & 0.199 & 0.317 & 0.318 & 0.168 &  - & - & - & - & - & - \\
OCT & \checkmark & 0.216
 & 0.199 & 0.105 & 0.209 & 0.187 & 0.102 &  - & - & - & - & - & - \\
OCT-global & & 0.232
 & 0.218  & 0.115 & 0.216 & 0.193 & 0.105 &  - & - & - & - & - & - \\
\hline
LLaVA-Pixel2Seq & & 0.156
 & 0.139 & 0.076  & 0.254 & 0.224 & 0.124 & 0.150 & 0.121 & 0.032 & 0.214 & 0.189 & 0.043 \\

LLaVA-Traj& & \textbf{0.126} & 0.142 & 0.073 &  0.201 & 0.191 & 0.103 & \textbf{0.133} & 0.130 & 0.031 & 0.191 & 0.167 & 0.041 \\

~\approach & & 0.136 & \textbf{0.106} & \textbf{0.062} & \textbf{0.194} & \textbf{0.157} & \textbf{0.090} & 0.135 & \textbf{0.108} &  \textbf{0.028} & \textbf{0.175} & \textbf{0.151} & \textbf{0.034} \\

\bottomrule
\end{tabular}
}
\caption{Comparison of VHP task with different baselines. We reported the performance on the validation split of Epic-Kitchen dataset. For the RBHP baselines, we also evaluate them on two unseen datasets, H2O and FPHA.}
\label{table:VHP}
\end{table*}
Following previous works \citep{liu2022jointhandmotioninteraction, ma2024diffip2ddiffusionbasedhandobjectinteraction} we use Average Displacement Error (ADE), Final Displacement Error (FDE) and Weighted Displacement Error (WDE) as metrics to evaluate VHP and RBHP tasks.

\textbf{Vanilla Hand Prediction.} For the VHP task, we choose Kalman Filter(KF) and Object-centric Transformer(OCT) \citep{liu2022jointhandmotioninteraction} as the baselines. Since OCT still requires the bounding box feature of the hand and object as input, to get a fairer comparison with other end-to-end methods, we implement a version without the requirement of the bounding box, which we call OCT-global.

\textbf{Reasoning-based Hand Prediction.} To evaluate~\approach's performance on the RBHP task, we perform baseline comparisons with several VLM-based methods. We describe these basleines below: 

\begin{itemize}[leftmargin=20pt,rightmargin=0pt]
\item \textbf{LLaVA-Traj.} Note that the hand trajectories are a sequence of pixel positions, we can represent them in text directly. In this case, we can directly fine-tune the LLaVA without any modification.
\item \textbf{LLaVA-Pixel2Seq.} An alternative approach to representing hand positions involves quantizing the image into discrete spatial bins \citep{chen2021pix2seq}, each corresponding to a unique token. We can extend the existing vocabulary with those discrete tokens. 
\item \rebuttalcolortext{\textbf{Language-conditioned Image-to-video Models. } We also compare our model to baselines of the language-conditioned image-to-video generation followed by hand-tracking. We use commercial state-of-the-art language-conditioned image-to-video systems such as LumaLabs~\citep{lumalabs2024dream}, Kling 1.5~\citep{klingai2024} and generate videos conditioned on the last observation frame and the language description. Following the hand label generation process in Sec.~\ref{sec:dataset generation process}, we track and extract the hand trajectories of the generated video.}
\end{itemize}

\subsection{Comparisons with Baselines}

We evaluate~\approach~on both the VHP task and the proposed RBHP task and report the results and comparisons with baselines in Table~\ref{table:VHP} and Table~\ref{tb:RBHP} respectively. All models except~\approach$^\dag$ are trained on VHP datasets.~\approach$^\dag$ is trained on all available datasets (Data Combo 5 in Table~\ref{tb:training_data}).

\textbf{VHP Task.} We evaluate all the baselines on the VHP datasets as described in section~\ref{sec:expdetails}. Here, the FPHA and H2O datasets serve as unseen datasets to test zero-shot generalization capabilities. Among all the VHP datasets,~\approach~outperforms both the task-specific methods as well as the VLM-based methods, which demonstrates its strong ability to produce plausible trajectories corresponding to how a real human hand would move given explicit instructions. We also find that~\approach~can generalize to completely unseen scenes (for example scenes from H2O and FPHA datasets), which demonstrates it can effectively leverage the world knowledge of the pre-trained VLM.

\textbf{RBHP Task.} For evaluations on the RBHP task shown in Table~\ref{tb:RBHP},~\approach~achieves state-of-the-art performance in all three metrics. This suggests that~\approach~ is able to reason based on implicit cues of the scene and be applied to complicated scenarios involving everyday natural language conversations. \rebuttalcolortext{However, we observe that LumaLabs~\citep{lumalabs2024dream} achieves the smallest ADE in the Ego4D RBHP benchmark but relatively higher FDE and WDE. This may be because the commercial text-conditioned image-to-video generation models have realistic video generation capabilities but cannot understand reasoning-based language prompts which is necessary for generating plausible videos maintaining temporal consistency. Since the training dataset compositions of these video models are not disclosed, there may also be some data leakage issues of the evaluation datasets in this paper being a part of their training corpora.}

\begin{wraptable}{r}{0.68\textwidth} 
\centering
\vspace*{-0.5cm}
\setlength\tabcolsep{2pt}
\begin{tabular}{c|ccc|ccc}
\toprule
&
\multicolumn{3}{c}{RBHP(Epic-Kitchen)} & \multicolumn{3}{c}{RBHP(Ego4D)}
\\

Approach & 
$\text {ADE} \downarrow$ &
$\text {FDE} \downarrow$ &
$\text {WDE} \downarrow$ &
$\text {ADE} \downarrow$ &
$\text {FDE} \downarrow$ &
$\text {WDE} \downarrow$
\\
\hline
Kling 1.5 & 0.311 & 0.358 & 0.197 & 0.277 & 0.411 & 0.184 \\
LumaLabs & 0.293 & 0.377 & 0.189 & \textbf{0.213} & 0.286 & 0.135 \\
\hline
% OCT-LLaVA &  &  &  & & & \\
LLaVA-Pixel2Seq & 0.277 & 0.248 & 0.137 & 0.312 & 0.287 & 0.143 \\
LLaVA-Traj & 0.196 & 0.187 & 0.101 & 0.381 & 0.353 & 0.178 \\
~\approach & 0.197 & 0.165 & 0.094 & 0.229 & 0.195 & 0.100 \\
~\approach$^\dag$ & \textbf{0.187} & \textbf{0.156} & \textbf{0.089} & 0.228 & \textbf{0.186} & \textbf{0.097} \\
\bottomrule
\end{tabular}
\caption{Comparison of~\approach~on the RBHP task with different baselines. \dag means fine-tuned on the RBHP dataset.}
\vspace*{-0.5cm}
\label{tb:RBHP}
\end{wraptable}

\subsection{Ablation Study}

In this section, we conduct a broad study of the different components of our model. All experiments in this section are evaluated on the RBHP task.

\textbf{Effects of Different Sources of Dataset.} In Table \ref{tb:training_data},
we show the contribution of each type of dataset to the performance of~\approach. LITA dataset denotes the different datasets for 5 additional tasks \citep{huang2024litalanguageinstructedtemporallocalization} described in Section~\ref{sec:expdetails} ranging from dense video captioning to reasoning about temporal localization. While increasing the scale of the VHP dataset (first two rows) can bring some improvement, we find that fine-tuning with the reasoning dataset (last two rows) can significantly boost the performance, even when fine-tuning with tasks that are not directly related to hand trajectory prediction. This demonstrates that~\approach~can leverage world knowledge learned by other tasks to reason about predicting plausible hand trajectories. 

\begin{table}[h]
\setlength{\tabcolsep}{9.3pt}
    \centering
\begin{tabular}{c|c|c|c|c|c|c|c}
\hline 
  & \multicolumn{2}{c|}{Epic-Kitchen} &   &  &  & \\
\cline{2-3} 
Data Combos& 55 & 100 & LITA data& RBHP data & ADE$\downarrow$ & FDE$\downarrow$ & WDE$\downarrow$ \\
\hline
1 & \checkmark & & & & 0.206 & 0.195 & 0.101 \\
2 & \checkmark & \checkmark & & & 0.197 & 0.165 & 0.094 \\
3 & \checkmark & \checkmark & \checkmark & & 0.199 & 0.163 & 0.094 \\
4 & \checkmark & \checkmark & \checkmark & \checkmark & \textbf{0.187} & \textbf{0.156} & \textbf{0.089} \\
\hline
\end{tabular}
    \caption{Analysis of the impact of training data on the performance of~\approach. We can see that performance increases with additional data of VHP (first two rows), even with datasets of other tasks (third row), but the highest gains come from the proposed RBHP dataset (last rows).}
    \label{tb:training_data}
\end{table}

\begin{wraptable}{r}{0.55\textwidth} 
    \centering
    \vspace*{-0cm}
\begin{tabular}{c|c|c|c}
\hline 
Num of Generations & ADE$\downarrow$ & FDE$\downarrow$ & WDE$\downarrow$ \\
\hline
1 & 0.187 & 0.156 & 0.089 \\
4 & 0.184 & 0.152 & 0.087 \\
8 & \textbf{0.182} & 0.151 & \textbf{0.086} \\
16 & \textbf{0.182} & \textbf{0.150} & \textbf{0.086} \\
\hline
\end{tabular}
    \caption{Analysis of test-time computations for~\approach~in the form of stochastic decoding with self-consistency~\citep{wang2023selfconsistency}.}
    \label{tab:tt}
\vspace*{-0.2cm}
\end{wraptable}

\textbf{Test-time Computation.} Recent works~\citep{snell2024scalingllmtesttimecompute, OpenAI2024} have shown that using more test-time computation is a critical step for LLMs to improve their performance, especially on reasoning tasks. Motivated by these, we also investigate if such properties can enhance the performance of~\approach~predictions. We report the performance using different numbers of generations during the stochastic decoding with self-consistency~\citep{wang2023selfconsistency} in Table \ref{tab:tt}. The main idea is to sample a diverse set of reasoning paths instead of just one and then select the most consistent output through marginalization. To obtain the self-consistency result in our context, we generate multiple answers for each inquiry and then average the predicted hand trajectory. We find that increasing the test-time computation in this form can robustly improve the performance of \approach as seen by the lower metrics from top to bottom in Table~\ref{tab:tt} .

\begin{figure}[t]
  \centering
  \includegraphics[width=1\textwidth,trim={0 3.5cm 0 0},clip]{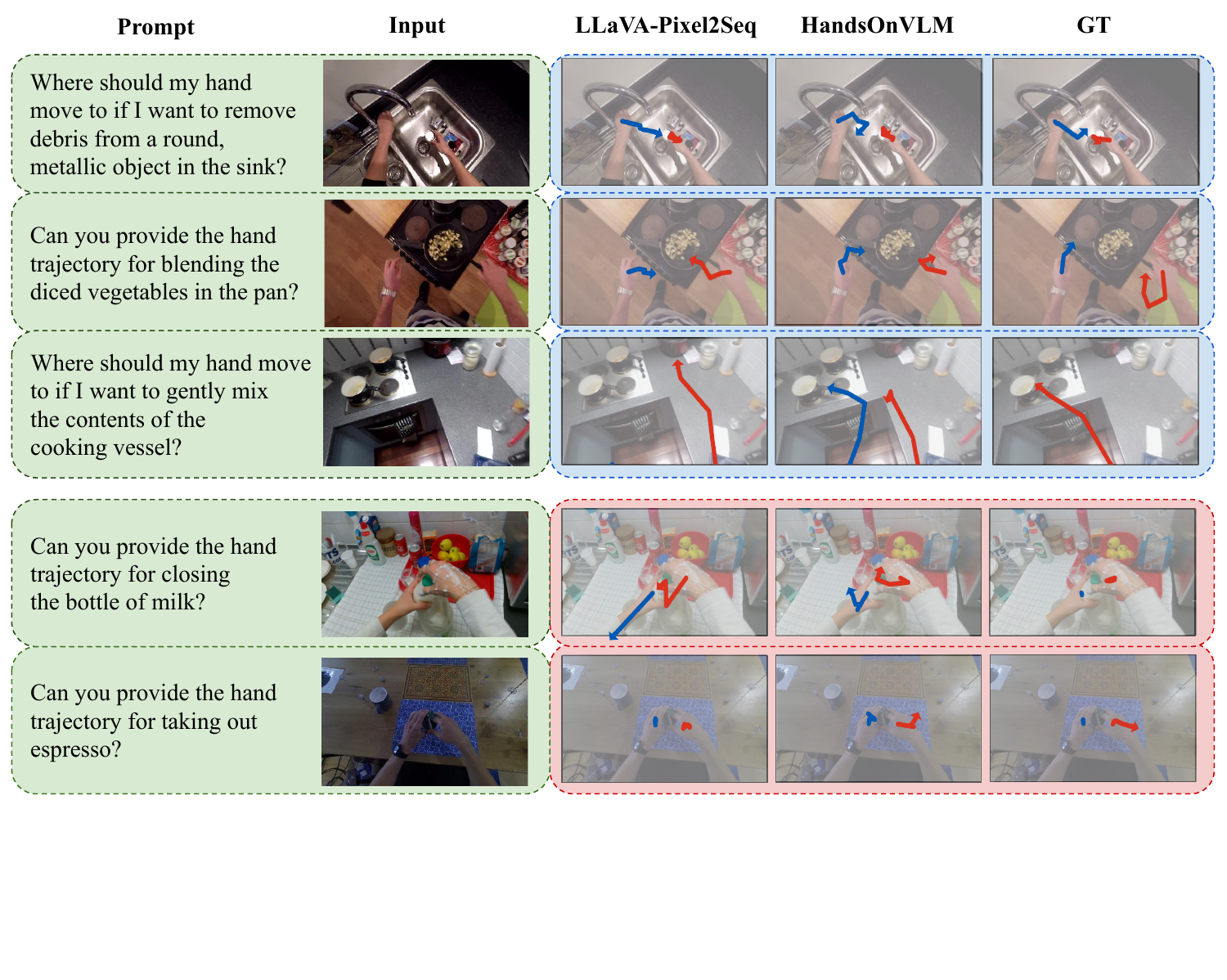}
    \vspace*{-0.8cm}
  \caption{Qualitative results for different samples from the validation split of our RBHP dataset (top in blue) and zero-shot evaluations on completely unseen datasets FPHA and H2O (bottom in pink). The left-hand trajectory is visualized in blue and the right-hand trajectory is in red. The arrows denote the direction of each trajectory. GT trajectories are provided for reference.}
  \label{fig:qualitative_visualization1}
  \vspace*{-0.6cm}
\end{figure}

\textbf{Effect of different Observation Frames.} \rebuttalcolortext{In Table \ref{tab:number of observation frames}, we investigate performance of our approach and the baselines when conditioned on just one observation frame instead of an observation video. Here we have four comparisons: OCT-last-im, OCT-global-last-im,~\approach-last-im,~\approach$^\dag$-last-im, which respectively correspond to versions of our baselines in Sec.~\ref{sec:baselines} but are only conditioned on the last frame of the input video context. We find that the results in this evaluation scenario are comparable to the setting where the context is a video, indicating that HandsOnVLM can flexibly be conditioned on just one image when a video context is not available.}

\begin{wraptable}{r}{0.75\textwidth} 
    \centering
    \vspace*{-0.5cm}
\begin{tabular}{c|c|c|c|c}
\hline 
Method & Num of Generations & ADE$\downarrow$ & FDE$\downarrow$ & WDE$\downarrow$ \\
\hline
\multirow{6}{*}{VHP} & OCT & 0.209 & 0.187 & 0.102 \\
& OCT-last-im & 0.213 & 0.191 & 0.104 \\
& OCT-global & 0.216 & 0.193 & 0.105 \\
& OCT-global-last-im & 0.212 & 0.189 & 0.103 \\
& ~\approach & \textbf{0.194} & \textbf{0.157} & \textbf{0.090} \\
& ~\approach-last-im & 0.197 & 0.165 & 0.094 \\
\hline
\multirow{4}{*}{RBHP} & ~\approach & 0.197 & 0.165 & 0.094 \\
& ~\approach-last-im & 0.197 & 0.163 & 0.093 \\
& ~\approach$^\dag$ & \textbf{0.187} & 0.156 & 0.089 \\
& ~\approach$^\dag$-last-im & \textbf{0.187} & \textbf{0.155} & \textbf{0.088} \\
\hline
\end{tabular}\vspace*{-0.2cm}
    \caption{Analysis of the number of observation frames during inference.}
    \label{tab:number of observation frames}
\vspace*{-0.3cm}
\end{wraptable}

\subsection{Qualitative Results}

In Fig.~\ref{fig:qualitative_visualization1} we show qualitative results for~\approach~and the strongest baseline LLaVA-Pixel2Seq.  The section above the horizontal line shows visualization from the validation split of RHBP datasets, while the section below  the line shows zero-shot results on scenes from completely unseen datasets. 

In the second row, we observe that~\approach~generates a trajectory where the left hand stably holds the pan while the right hand performs the blending action. In contrast, LLaVA-Pixel2Seq fails to correctly depict holding the pan. The third row results demonstrates~\approach's ability to reason about multi-modal solutions for the same task. While the ground truth shows the right hand moving the pot,~\approach~chooses to use the left hand to execute the same action, illustrating its multi-modal reasoning ability capability.

\section{Conclusion}
\noindent\textbf{Summary.} In this work, we propose~\approach, a novel video-based VLM to predict hand motion from ego-centric videos. We also proposed two tasks, Vanilla Hand Prediction(VHP) task and Reasoning-based Hand Prediction(RBHP) task to benchmark the hand motion prediction as well as the reasoning ability. We demonstrate its effectiveness through extensive quantitative and qualitative results. We believe this research represents a promising initial step towards integrating egocentric hand-object video understanding with the powerful capabilities of VLMs.

\noindent\textbf{Limitations.} While we enabled hand-trajectory prediction from colloquial language instructions, the quality of our predictions are bottle-necked by the limitations of ground-truth hand location extraction from videos, the models for which often fail when the hand is occluded or moving too fast. In addition, the 2D locations of hand we predict are not rich enough for directly being adapted for downstream applications like robotics and augmented reality. 

\noindent\textbf{Future Work.} An interesting direction of future work would be to predict trajectories of full hand meshes in the future including orientation and articulation and also include depth in the predictions. Another exciting direction would be to adapt our model for long-horizon predictions for activities like ``making coffee" which would consist of several steps and require reasoning over an extended period. Since video clips on the web have significant camera motion over time, a viable strategy for this could be chaining the model sequentially for different sub-tasks. 

\section*{Acknowledgments}
We thank Gaurav Parmar and Jinkun Cao for feedback on the paper, and thank Yufei Ye, Ruihan Yang, Unnat Jain, Mohan Kumar Srirama, Shubham Tulsiani, and many others from CMU and UCSD for helpful discussions. This work used Bridges-2 at Pittsburgh Supercomputing Center from the Advanced Cyberinfrastructure Coordination Ecosystem: Services \& Support (ACCESS) program, which is supported by National Science Foundation grants 2138259, 2138286, 2138307, 2137603, and 2138296.

\bibliography{iclr2025_conference}
\bibliographystyle{iclr2025_conference}

\newpage
\appendix
\section{Appendix}
Here we provide additional details of the model implementation, dataset curation, and more qualitative results.

\subsection{Dataset Details}
\label{gtgeneration}

\textbf{Statistics.} Table \ref{tb:Dataset Statistics} shows the statistics of all datasets used in our tasks. Note that H2O, FPHA and Ego4D are only used for zero-shot evaluation so there are no training samples.
\begin{table}[htbp!]
\setlength{\tabcolsep}{9.3pt}
    \centering
\begin{tabular}{c|c|c|c}
\hline 
Task & Dataset & Training Samples & Validation Samples \\
\hline 
\multirow{4}{*}{VHP} & Epic-Kitchen-55 & 8523 & 1894 \\
& Epic-Kitchen-100 & 24148 & 3513 \\
& H2O & - & 503 \\
& FPHA & - & 501 \\
\hline 
\multirow{2}{*}{RBHP} & Epic-Kitchen-100 & 4018 & 3513 \\
& \rebuttalcolortext{Ego4D} & - & \rebuttalcolortext{8673} \\
\hline
\end{tabular}
    \caption{Data Statistics of VHP and RBHP task.}
    \label{tb:Dataset Statistics}
\end{table}

\subsection{Training Approach Details}

Here we provide the illustrations of the training pipeline and the inference pipeline.

\begin{figure}[h]
  \centering
  \includegraphics[width=0.9\textwidth,trim={0 2.3cm 0 0},clip]{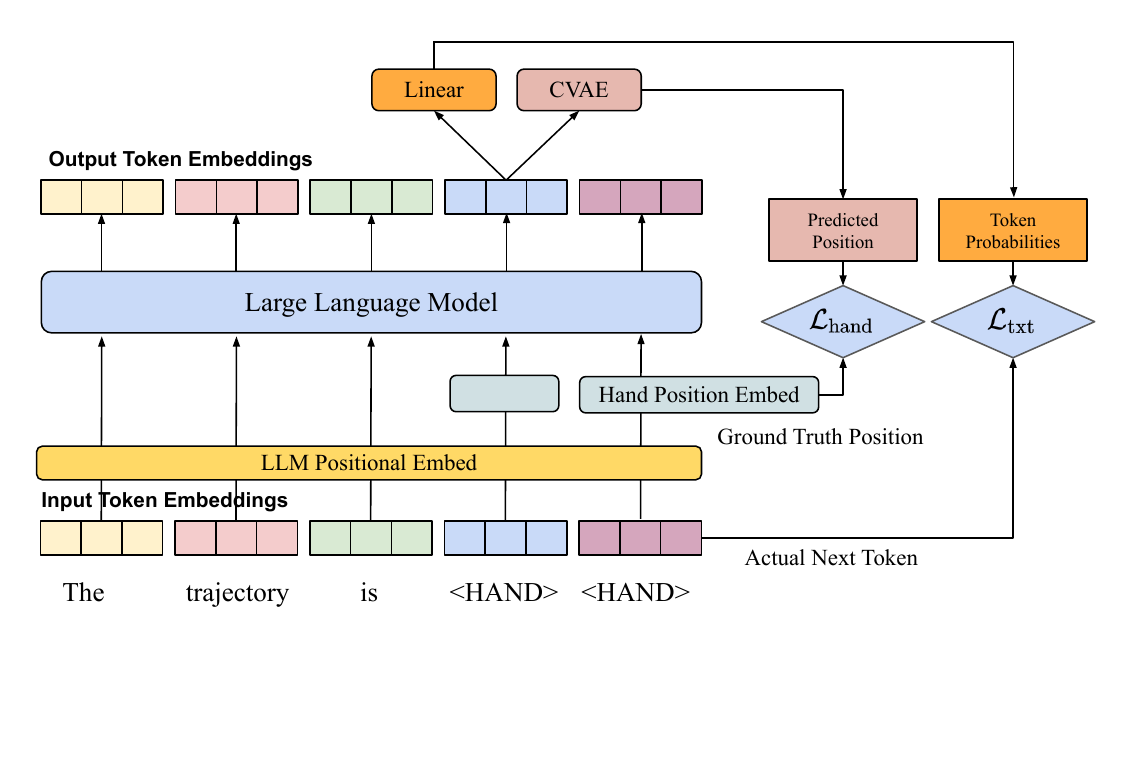}
  \caption{Illustration of training pipeline.}
  \label{fig: train piepeline}
\end{figure}

\begin{figure}[h]
  \centering
  \includegraphics[width=0.9\textwidth,trim={0 1.8cm 0 0},clip]{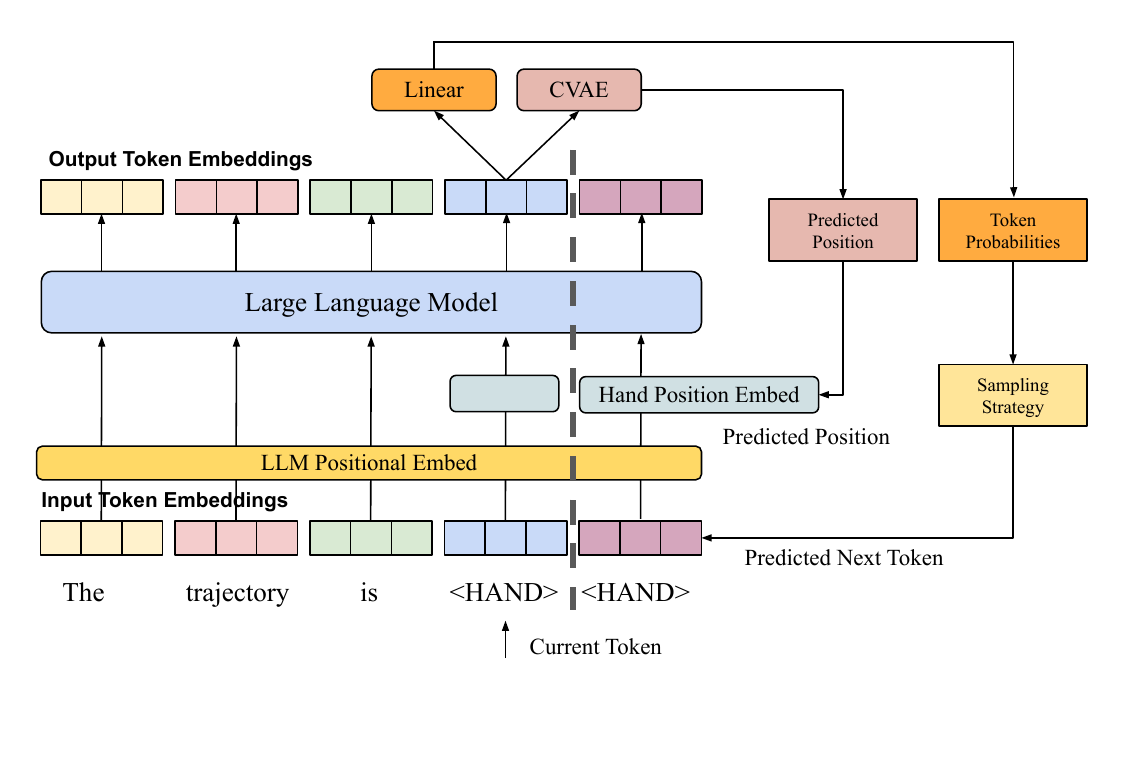}
  \caption{Illustration of inference pipeline.}
  \label{fig: inference piepeline}
\end{figure}

\subsection{Other Ablation Studies}

\textbf{Scaling Model Improves the Prediction.} To evaluate the scaling ability of our model, we use LLaVA-V1.3-7B and LLaVA-V1.3-13B as the LLM backbone of our model. We refer them as~\approach-7B and~\approach-13B. We show the performance of both models in Fig. \ref{tab:llm model size}.

\textbf{Zero-shot Chain-of-thought.} We also conduct an ablation study on the zero-shot chain-of-thought \citep{wei2022chain, kojima2022zeroshotcot} prompting, as shown in Fig. \ref{tab:chain of thought}. We add ``Let's think step by step" in the front of the answer generated in the inference stage. Contrary to our expectations, this approach yielded poorer results. This unexpected outcome may be attributed to the limited diversity of our datasets.

\begin{figure}[h!]
    \centering
    \begin{minipage}{0.48\textwidth}
        \centering
    \begin{tabular}{c|c|c}
        \hline 
        Approach & ADE$\downarrow$ & FDE$\downarrow$ \\
        \hline
        \approach-7B & 0.197 & 0.165 \\
        \approach-13B & \textbf{0.183} & \textbf{0.149} \\
        \hline
        \end{tabular}
        \caption{Ablation study on the LLM backbone size. We evaluate them on the RBHP task.}
        \label{tab:llm model size}
    \end{minipage}%
    \hfill
    \begin{minipage}{0.48\textwidth}
        \centering
        \begin{tabular}{c|c|c}
        \hline 
        Reasoning Method & ADE$\downarrow$ & FDE$\downarrow$ \\
        \hline
        Direct Answer & \textbf{0.197} & \textbf{0.165} \\
        Chain-of-Thought & 0.220 & 0.191 \\
        \hline
        \end{tabular}       \caption{Comparison of direct answer and chain-of-thought reasoning methods.}
        \label{tab:chain of thought}
    \end{minipage}
\end{figure}

\subsection{More Visualizations}

\begin{figure}[htbp!]
  \begin{minipage}{0.31\linewidth}
    \centering
    \includegraphics[width=\linewidth,trim={0 0cm 0 0},clip]{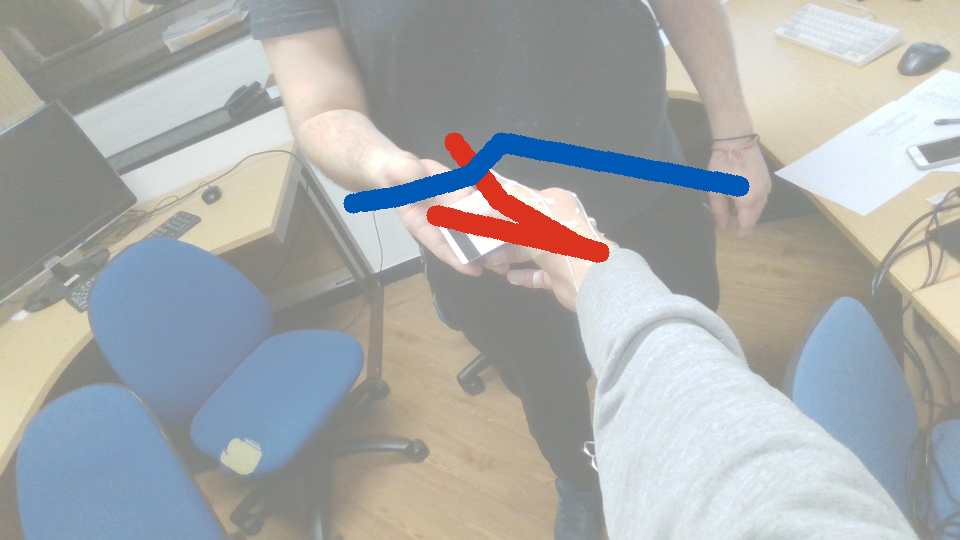}
  \end{minipage}
  \hfill
  \begin{minipage}{0.31\linewidth}
    \centering
    \includegraphics[width=\linewidth,trim={0 0cm 0 0},clip]{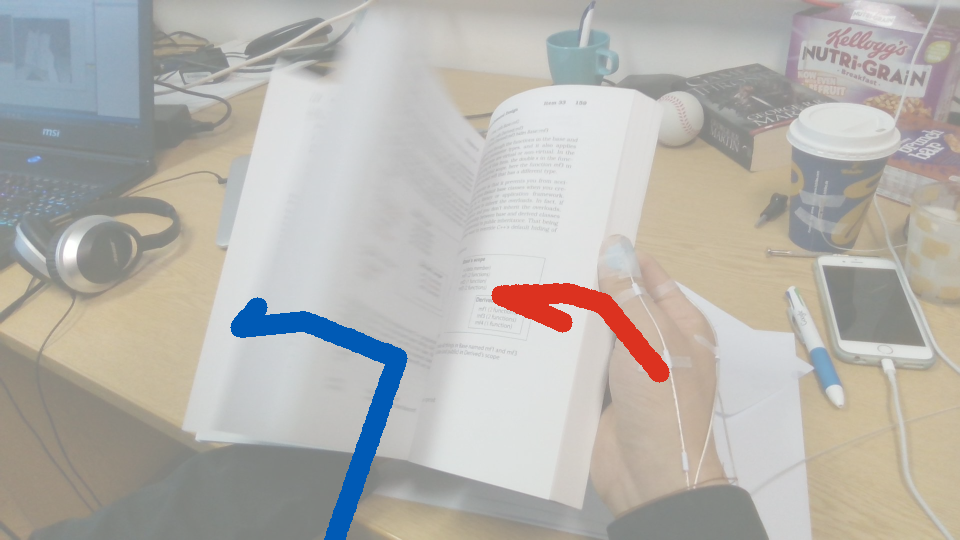}
  \end{minipage}
  \hfill
  \begin{minipage}{0.31\linewidth}
    \centering
    \includegraphics[width=\linewidth,trim={0 0cm 0 0},clip]{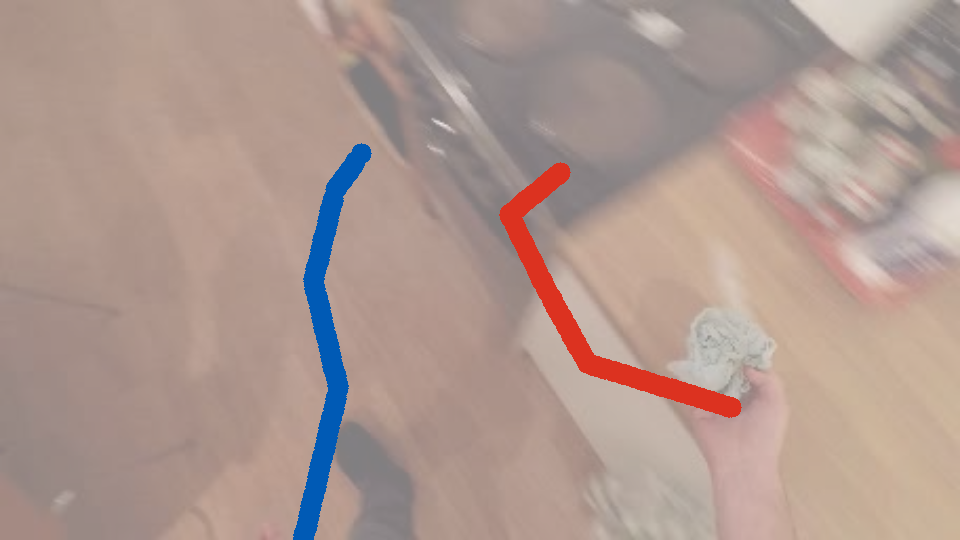}
  \end{minipage}
  \caption{Failure cases of the model: (left) multiple hands in the video, (middle) occlusions, and (right) the target trash can is out of view.}
    \label{fig:failure}
\end{figure}

\textbf{Failure Cases.}
We show some failure cases in Fig. \ref{fig:failure}. We observe failures when (1) there are other people's hands in the context video/image, (2) the hands are occluded by objects, and (3) the target object in the instruction is not visible in the context frame.

\begin{figure}[t]
  \centering
  \includegraphics[width=1\textwidth,trim={0 8cm 0 0},clip]{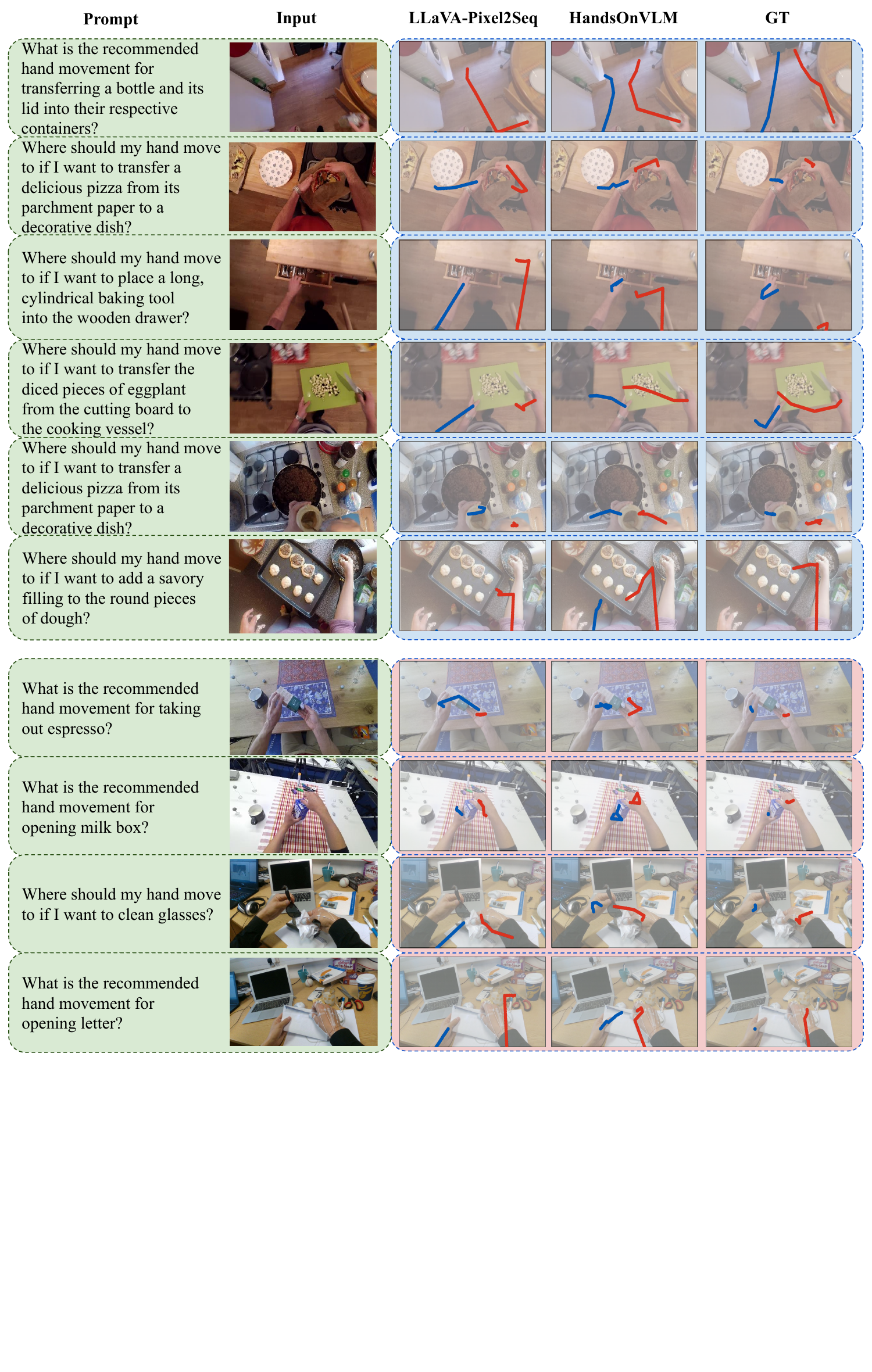}
  \caption{More Qualitative results for different samples from the validation split of our RBHP dataset (top in blue) and zero-shot evaluations on completely unseen datasets FPHA and H2O (bottom in pink). GT trajectories are provided for reference.}
  \label{fig:qualitative_visualization2}
\end{figure}

\textbf{More Qualitative Results.} We provide more visualizations in Fig. \ref{fig:qualitative_visualization2}.

\subsection{Prompt for VHP and RBHP Dataset Generation}
\label{sec: prompts}
We provide the GPT4 prompts for the RBHP dataset generation pipeline mentioned in Section \ref{sec:reasoning data annotation pipeline} in Table \ref{tab:llm-prompt} and Table \ref{tab:hand-trajectory-prompt}.

\begin{table}[p!]
\centering
\begin{tabular}{|p{\textwidth}|}
\hline
\textbf{GPT4 Prompt for Action-aware Image Description} \\
\hline
You are a system generating descriptions for ego-centric human images. Human is doing household activities. \\ \\
Provided with an image and a action narration of what is happening next, such as ``use the scissor", you will describe the main item that you see in the image, giving details but staying concise. \\ \\
You can describe unambiguously what the item is, its color or relative position if clearly identifiable. \\
You should also give out a overall description of the scene, the environment where the action is taking place. \\
\hline
\end{tabular}
\caption{GPT4 prompt for action-aware image description.}
\label{tab:llm-prompt}
\end{table}

\begin{table}[p!]
\centering
\begin{tabular}{|p{\textwidth}|}
\hline
\textbf{GPT4 Prompt for Implicit Action Generation} \\
\hline
You are tasked with creating specific, indirect questions and instructions that human could use to identify and interact with objects based on their names or detailed descriptions provided by users. \\ \\
You will be given an action phrase which the human is going to do next, such as ``use the scissor". \\ \\
Based on the descriptions, you must formulate responses that precisely hint at the action phrase without naming it directly. The aim is to enable the agent to deduce the correct action through these indirect cues, enhancing its ability to understand and execute tasks involving the object. \\ \\
Please format your generated response as a hand trajectory question, some templates are provided below for reference: \\
``Where should my hand move to if I want to \{implicit description\}" \\
``Can you provide the hand trajectory for \{implicit description\}?" \\
``What is the recommended hand movement for \{implicit description\}?" \\
\hline
\end{tabular}
\caption{GPT4 prompt for implicit action generation.}
\label{tab:hand-trajectory-prompt}
\end{table}

\begin{table}[p!]
\centering
\begin{tabular}{|p{\textwidth}|}
\hline
\textbf{Question Templates to Build VHP Datasets.} \\
\hline
``Can you provide the hand trajectory?"\\
``What is the recommended hand movement?"\\
``What is the future hand trajectory in this video?"\\
``What is the predicted hand trajectory given current observations?"\\

``Where should my hand move to if I want to \{explicit action\}?"\\
``Can you provide the hand trajectory for \{explicit action\}?"\\
``What is the recommended hand movement for \{explicit action\}?"\\
\hline
\end{tabular}
\caption{Question Templates to build VHP datasets.}
\label{tab:llm-prompt-question}
\end{table}

\begin{table}[p!]
\centering
\begin{tabular}{|p{\textwidth}|}
\hline
\textbf{Answer Templates to build VHP and RBHP datasets.} \\
\hline
``Sure! Here is the hand trajectory \{hand token sequence\}." \\
``Based on the video, the hand trajectory is as follows: \{hand token sequence\}." \\
``The predicted hand trajectory is as follows: \{hand token sequence\}." \\
``Certainly! The hand trajectory for \{action instruction\} is as follows: \{hand token sequence\}." \\
``To \{action instruction\}, the recommended hand trajectory is: \{hand token sequence\}." \\
\hline
\end{tabular}
\caption{Answer Templates to build VHP and RBHP datasets.}
\label{tab:llm-prompt-answer}
\end{table}

\end{document}

%% file: math_commands.tex
\usepackage{amsmath,amsfonts,bm}

\def\eqref#1{equation~\ref{#1}}
\def\1{\bm{1}}

\DeclareMathAlphabet{\mathsfit}{\encodingdefault}{\sfdefault}{m}{sl}
\SetMathAlphabet{\mathsfit}{bold}{\encodingdefault}{\sfdefault}{bx}{n}